# Low-cost and high-performance data augmentation for deep-learning-based skin lesion classification

*Shuwei Shen[a,#], Mengjuan Xu[b,#], Fan Zhang[b], Pengfei Shao[b], Honghong Liu[a], Liang Xu[b], Chi Zhang[a], Peng Liu[b], Zhihong Zhang[a,\*], Peng Yao[b,\*], Ronald X. Xu[b,c,\*]*

[a]*First Affiliated Hospital, University of Science and Technology of China, Hefei 230031, China*
[b]*Department of Precision Machinery and Precision Instrumentation, University of Science and Technology of China, Hefei 230026, China*
[c]*Department of Biomedical Engineering, The Ohio State University, Columbus, OH 43210, USA*

*Abstract*

Although deep convolutional neural networks (DCNNs) have achieved significant accuracy in skin lesion classification comparable or even superior to those of dermatologists, practical implementation of these models for skin cancer screening in low resource settings is hindered by their limitations in computational cost and training dataset. To overcome these limitations, we propose a low-cost and high-performance data augmentation strategy that includes two consecutive stages of augmentation search and network search. At the augmentation search stage, the augmentation strategy is optimized in the search space of Low-Cost-Augment (LCA) under the criteria of balanced accuracy (BACC) with 5-fold cross validation. At the network search stage, the DCNNs are fine-tuned with the full training set in order to select the model with the highest BACC. The efficiency of the proposed data augmentation strategy is verified on the HAM10000 dataset using EfficientNets as a baseline. With the proposed strategy, we are able to reduce the search space to 60 and achieve a high BACC of 0.853 by using a single DCNN model without external database, suitable to be implemented in mobile devices for DCNN-based skin lesion detection in low resource settings.

## 1. Introduction

Skin diseases represent one of the most common health problems in the world(Seth, Cheldize, Brown, & Freeman, 2017) that affect patient's quality life, induce significant socioeconomic burden to the society, and even lead to increased morbidity and mortality(Amarathunga, Ellawala, Abeysekara, & Amalraj, 2015; "Cancer Facts & Figures, American Cancer Society, [online]," 2020). Skin cancer is a family of skin diseases caused by neoplastic growth of skin cells in the epidermis and can be classified into two major categories of non-melanoma and melanoma(Linares, Zakaria, & Nizran, 2015). Non-melanoma skin cancer (NMSC) accounts for 98% of all the skin cancers, and their treatment places a significant burden to the healthcare systems (Machlin, Carper, & Kashihara, 2011). Melanoma accounts for only 2% of all the skin cancers but causes the most skin cancer deaths (Linares, et al., 2015). Early detection and prompt treatment of skin lesions can greatly improve quality of life and reduce melanoma mortality for patients. Previous study has revealed an elevated 5-year survival of 99% for early detected melanoma, in comparison with that of ~18% with late diagnosis(Akar, Marques, Andrews, & Furht, 2019). Current standard of care for skin cancer diagnosis is based on visual inspection of lesion size, shape, color, and location (Habif, Chapman, Dinulos, & Zug, 2017). Although using a dermoscope helps improving the diagnostic accuracy (Rosendahl, Tschandl, Cameron, & Kittler, 2011), visual inspection represents a subjective method for skin cancer detection, and its accuracy heavily dependent on the examiner's experience(Celebi, et al., 2007) . Due to the global shortage in the experienced dermatologists, patents in rural communities and low resource settings have experienced the significant delay in detection and treatment of skin cancer as well as the higher morbidity and mortality compared with other areas (Seth, et al., 2017).

To address the shortage of dermatology specialists and improve the accuracy for skin cancer classification, various artificial intelligent (AI) diagnostic technologies have been explored (Abuzaghleh, Barkana, & Faezipour, 2015; Smaoui & Bessassi, 2013). Since the first report in 1987(Cascinelli, Ferrario, Tonelli, & Leo, 1987), traditional machine learning techniques have been applied to help dermatologists in faster data process and more reliable diagnosis(Oliveira, et al., 2016; Zakeri & Hokmabadi, 2018). Machine learning algorithms, such as SVM, weighted K-NN, and decision tree classifier, have already achieved the classification rates of 80%, 96.36%, 89.5%, respectively (Çataloluk & Kesler, 2012; Taufiq, Hameed, Anjum, & Hameed, 2017; Victor, Ghalib, & Systems, 2017). However, the performance of these methods in multiclass classification is limited by many deficiencies such as excessive adjustment (i.e., over-fitting). With recent advances in computing technology, deep convolutional neural network (DCNN) has been introduced into the skin disease diagnosis(Marchetti, et al., 2018) and has achieved the encouraging diagnostic accuracies better than 90%, comparable or even superior to those of dermatologists (Ech-Cherif, Misbhauddin, & Ech-Cherif, 2019; Esteva, et al., 2017; Haenssle, et al., 2018; Han, et al., 2018).

Although DCNN has been integrated with several mobile dermoscopic devices for intelligent classification of skin lesions (Ech-Cherif, et al., 2019; Hameed, et al., 2020; Sallam & Alawi, 2019), rural deployment of DCNN classifiers for skin cancer screening is hindered by multiple technical challenges in computational cost, portability, and reliability. It has been revealed that DCNN classifiers tend to work well when they are trained on large datasets acquired from actual clinical cases(Deng, et al., 2009). However, most of the publically accessible skin lesion imaging depositories, such as PH[2](Mendonça, Ferreira, Marques, Marcal, & Rozeira, 2013) (160 nevi and 40 melanomas), ISIC 2017(13768 images, and the number of AKIEC is only 2), ViDIR series(Philipp Tschandl, et al., 2019) (<4,000 images with partial pathologic verification), and HAM10000 (10015 images in seven skin disease categories)(P. Tschandl, Rosendahl, & Kittler, 2018), have either a relatively small data size or an uneven distribution of disease types, limiting the achievable performance of the trained classifiers (Codella, et al., 2018; Dreiseitl, Binder, Hable, & Kittler, 2009; Mendonça, et al., 2013). Similarly, lack of large scale, reliable and balanced clinical dataset presents a common barrier for development and deployment of high performance DCNN models in many other clinical disciplines.

Ensembling multiple deep learning (DL) models and data augmentation are commonly used methods to overcome the aforementioned limitations of the available clinical datasets. Model ensembling is a process that aggregates the predictions of multiple diverse models in one final prediction, where the diverse models are trained using different strategies(Elkan, 2013). This approach improves the prediction performance by reducing the generalization error of the prediction, as evidenced by the observation that the balanced multiclass accuracy of the ensembling models is generally higher than that of the single models in the ISIC 2018 challenge(Nozdryn-Plotnicki, Yap, & Yolland, 2018). However, the computational cost of the ensembling models is generally high and the ceiling effect prevents their further improvement of diagnostic accuracy. Therefore, it is necessary to improve the performance of individual DCNN models and increase both "amount" and "diversity" of the existing dataset by applying various data augmentation strategies that are typically dataset-specific (Cubuk, Zoph, Mane, Vasudevan, & Le, 2019; Krizhevsky, Sutskever, & Hinton, 2012). The conventional data augmentation strategies include

---

[#]*These authors contributed equally.*
*Corresponding author.*
E-mail address: swshen@ustc.edu.cn (S. Shen), xumj@mail.ustc.edu.cn (M.Xu), zxt1002@mail.ustc.edu.cn (F. Zhang), spf@ustc.edu.cn(P. Shao), xiaowandou007@163.com (H, Liu), xul666@mail.ustc.edu.cn (L. Xu), zcwill@163.com (C. Zhang), lpeng01@ ustc.edu.cn (P. Liu), zzhzqr@126.com (Z. Zhang), yaopeng@ustc.edu.cn (P. Yao), xu.202@osu.edu (R. Xu)

scaling, translation, rotation, random cropping, image mirroring, and color change (Cubuk, et al., 2019; Sato, Nishimura, & Yokoi, 2015; Wan, Zeiler, Zhang, Le Cun, & Fergus, 2013). Recently, the emerging automatic data augmentation techniques, such as AutoAugment(Cubuk, et al., 2019), have shown a certain superiority over conventional data augmentation strategies (Ho, Liang, Chen, Stoica, & Abbeel, 2019; Zoph, et al., 2019). Nevertheless, the high computational cost prevents the practical implementation of the ensembling models and data augmentation strategies in low resource settings (Cubuk, Zoph, Shlens, & Le, 2020).

We propose a low-cost and high-performance data augmentation strategy suitable to be implemented in mobile devices for AI-based skin lesion detection. The proposed strategy includes two consecutive stages of augmentation search and network search. At the stage of augmentation search, the best augmentation strategy is searched in the space of Low-Cost-Augment (LCA) under the criteria of balanced accuracy (BACC) with 5-fold cross validation. At the stage of network search, the DL models are fine-tuned with the full training set and the model with the highest BACC is selected. In this paper, the efficiency of such a data augmentation strategy is validated on the HAM10000 dataset using the EfficientNet model as a baseline. EfficientNet is a group of lightweight convolutional network models achieving state-of-the-art accuracy with an order of magnitude fewer parameters and FLOPS over ImageNet(Tan & Le, 2019). Our research contributions can be summarized as follows:

1, Our study implies that the best combination of the argumentation strategy and the network should be re-searched for different datasets.

2, By training the EfficientNet model using the proposed data augmentation strategy on the HAM10000 dataset, we have achieved an BACC level of 0.853, ranking top of the ISIC 2018 challenge in the channel of single-model and no-external-database.

3, With the combination of the proposed DL model, we have achieved a high BACC at a low computation cost, making the technique readily implementable in mobile dermoscopic devices for skin cancer screening in rural communities.

The proposed DL strategy can be expanded to other clinical disciplines for automatic screening and accurate detection of diseases in areas of limited resources. All the source code used in this paper is available in our public repository[1].

## 2. Methodology

### 2.1 Training and search strategies

Figure 1 illustrates the two-stage approach in search for the best combination of the augmentation and the network strategies for a specific clinical dataset. Prior to the search tasks, a novel data augmentation search space is defined and the original dataset is divided into a training set and a test set. The training set is further divided into training and validation subsets. At stage one, the best augmentation strategy for the original dataset in the proposed data augmentation space is searched using 5-fold cross-validation. The search is based on the training set and the candidate DCNN models, where training and validation subset are randomly updated in different folds and the average of five best BACCs in the validation subsets is adopted as the screening criterion. At stage two, the DCNNs are refined by applying the best augmentation strategy using the full training set. After these two stages, the DCNN model with the best BACC on the test dataset and the best augmentation strategy will be preferentially recommended for subsequent clinical classification tasks.

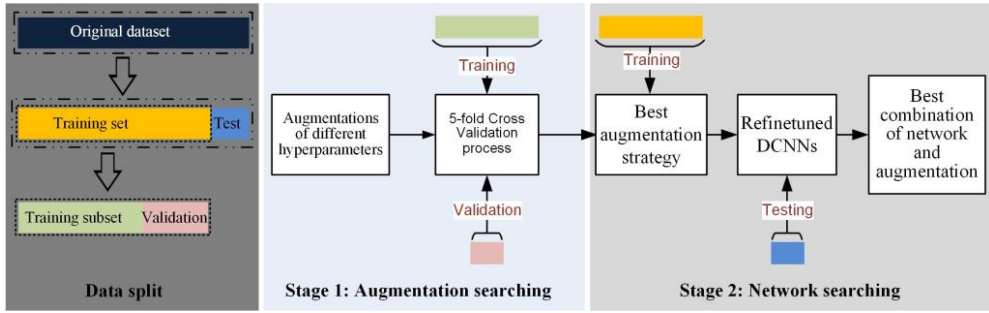

**Fig. 1 - Work flow for the two-stage approach to search for the best combination of network and augmentation strategy.**

In order to reduce the computational cost for the two-stage searching method, we propose a novel LCA augmentation search space that can be used extensively across different datasets for significant reduction of the search space. The proposed augmentation policies include not only the flip and scale changes and the AutoAugment strategies (Ho, et al., 2019; Zoph, et al., 2019), but also the randomly added Gaussian noise(DeVries & Taylor, 2017) and the color tone shift(Wu, Yan, Shan, Dang, & Sun, 2015). As shown in Table 1, the LCA search space is defined as an unordered set of 12 sub-policies, and one of the sub-policies is randomly selected and executed at the associated probability. Each sub-policy contains a color change operation and a randomly paired geometric change operation. Moreover, color operations vary from different sub-policies, whereas all geometric operations appear at the same frequency. We treat the augmentation search process as a discrete optimization problem(Cubuk, et al., 2019), which greatly reduces the requirements for computational resources and thereby shortens the search time. In detail, the probability for the randomly paired operations is set as a ladder parameter in all the 5 folds cross-validation tests, and the magnitude of each operation is randomly determined within the specified range as in Table X1. This search process introduces a notion of stochasticity into the augmentation policy, therefore further enhances the robustness of the augmentation strategy.

**Table 1 - List of all the sub-policies in LCA-based augmentation.**

| Policy | Operation 1 | Operation 2 | Probability | Policy | Operation 1 | Operation 2 | Probability |
|---|---|---|---|---|---|---|---|
| Sub-policy1 | Sample_Pairing | Rotate | [0, 1] | Sub-policy7 | Sharpness | Rotate | [0, 1] |
| Sub-policy2 | Gaussian_noise | Flip | [0, 1] | Sub-policy8 | Color_shift | Scale | [0, 1] |
| Sub-policy3 | SolarizeAdd | Cutout | [0, 1] | Sub-policy9 | Equalize_YUV | ShearX | [0, 1] |
| Sub-policy4 | Color | ShearX | [0, 1] | Sub-policy10 | Posterize | ShearY | [0, 1] |
| Sub-policy5 | Contrast | ShearY | [0, 1] | Sub-policy11 | AutoContrast | Flip | [0, 1] |
| Sub-policy6 | Brightness | Scale | [0, 1] | Sub-policy12 | Equalize | Cutout | [0, 1] |

### 2.2 Datasets and DCNNs

The efficiency of the LCA-based data augmentation strategy is validated on a publically accessible HAM10000 dataset and representative DCNNs. HAM10000 dataset contains 10015 skin lesion images. Out of them, 6705 are melanocytic nevi (NV); 1113 are melanoma (MEL); 1099 are benign keratosis-like lesion (BKL); 514 are basal cell carcinoma (BCC); 327 are Actinic keratosis/Bowen's diseases (AKIEC); 142 are vascular(VASC) and 115 are dermatofibroma (DF) (P. Tschandl, et al., 2018). Most of the HAM10000 images have the target lesions located at the center, and 53.30% of them are pathologically verified. HAM10000 is a publically accessible dataset for the 2018 skin lesion analysis challenge organized by International Skin Imaging Collaboration (ISIC) (https://www.isic-archive.com). Figure 2 shows a representative image from this dataset and a variety of images after applying different enhancement sub-strategies. The HAM10000 dataset is selected for testing our data augmentation strategy since its performance outcome can be easily compared with those of many other strategies published on the leaderboard.

---

[1] github.com/Shuwrood-SSW/Low-cost-and-high-performance-data-augmentation-for-deep-learning-based-skin-lesion-classification

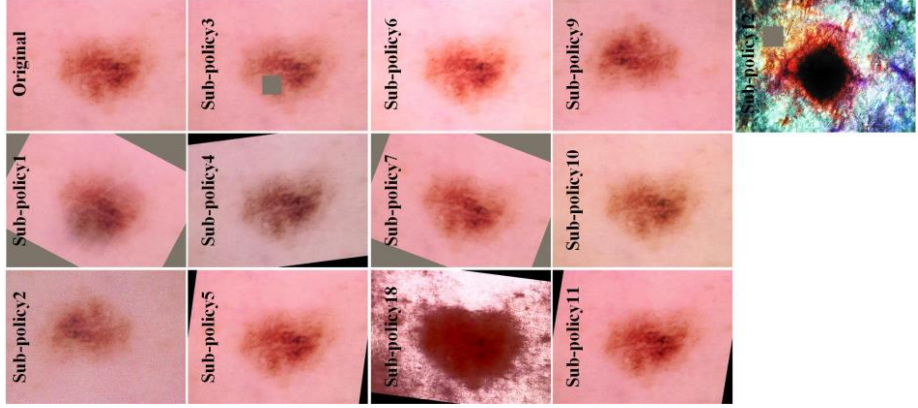

**Fig. 2 - Representations of augmentation effects of different sub-strategies in the probability of 0.5.**

In terms of the DCNN models, we have evaluated the pre-trained architectures and found that finely tuning a model trained on ImageNet performed significantly better than that trained from scratch. Previous studies also show better performance by using more recent architectures (Gessert, et al., 2018). After comparing with several classic models such as Inception(Szegedy, Vanhoucke, Ioffe, Shlens, & Wojna, 2016), ResNet(He, Zhang, Ren, & Sun, 2016), PolyNet(Zhang, Li, Change Loy, & Lin, 2017), and Densenet (Huang, Liu, Van Der Maaten, & Weinberger, 2017), we finally select the state-of-the-art EfficientNet model due to its better ImageNet accuracy that requires fewer parameters and FLOPS (Tan & Le, 2019).

*2.3 Optimization and validation of the training strategy*

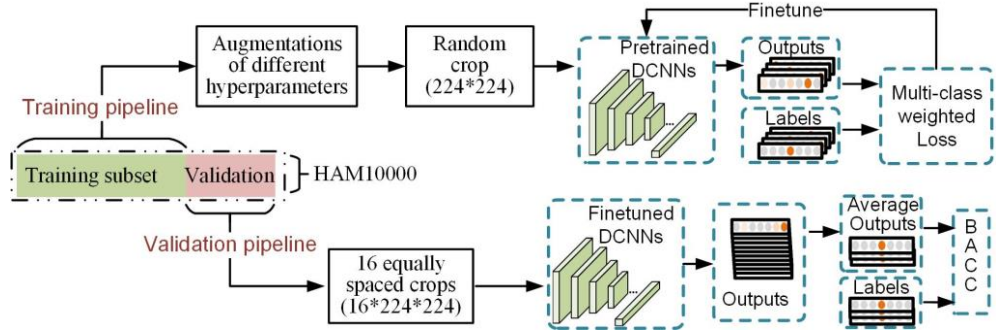

**Fig. 3 - Implementing details in one of 5-fold cross-validation**

Considering the important role that a training strategy plays in the final performance of DCNNs, we optimize the training strategy by implementing a 5-fold cross-validation procedure where the training set (HAM10000) is split into the training and the validation subsets following a ratio of 4:1 in each fold, as illustrated in figure 3. The subset separation procedure ensures that the same lesion does not occur in both the training and the validation subsets and that the compositions of subsets vary randomly from different folds (Gessert, et al., 2018). Especially, the strategy is finely tuned on the full HAM10000 dataset by applying the previously identified augmentation strategy at the stage of network searching, and the predicted classification results on the validation dataset have been uploaded to the ISIC challenge website for verification. Each image in the training set has an initial size of 600 x 450, and a randomly selected sub-strategy is executed following the defined probability. In this regard, the ladder probabilities are set as [0.1, 0.3, 0.5, 0.7, 0.9]. The image of 224 x 224 is randomly cropped(Howard, 2013) from the augmented images and subsequently plug into the pre-trained DCNNs with the modified output dimension of 7. In terms of the loss function, the following standard cross-entropy loss is used as a basis:

$$\mathcal{L} = -\sum_{i=1}^{C} p_i * \log \hat{p}_i \qquad (1)$$

where $p_i$ is the ground-truth label of class $i$, $\hat{p}_i$ is the softmax normalized model output and C is the number of classes.

Since the seven classes in the original HAM10000 dataset are highly imbalanced, the multi-class weighted loss is implemented by adding an enhanced weight on the underrepresented classes, such as the highly underrepresented DF and VASC, to improve the overall performance of the trained DCNNs(Gessert, et al., 2018). The multi-class weighted loss is updated by multiplying class-equilibrium matrix with the standard cross-entropy loss function, where the enhanced weight $w_i$ for class $i$ corresponds to the inverse normalized class frequencies (equation 2). The multi-class weighted loss is therefore updated as in equation 3:

$$w_i = N/n_i \qquad (2)$$
$$\mathcal{L} = -\sum_{i=1}^{C} w_i * p_i * \log \hat{p}_i \qquad (3)$$

where N is the total number of samples and $n_i$ is the number of samples for class *i*.

The selection of other hyperparameters is more straight forward. First, we choose a starting learning rate of lr = 0.001 and reduce it with a factor of λ =1/10 after 20 epochs. Then, we continue reducing the learning rate with the same factor at every 10 epochs, and stop the optimization after 70 epochs. Finally, we select the best performing Adam as the optimizer for all the models. Considering that the same number of Graphics Processing Unit (GPU) carded are located for parallel searching tasks and that the feature map size of DCNNs increases proportionally with their parameters, using a uniform batch-size may result in insufficient or waste of the computational resources. Therefore, we set the batch-sizes for different models as model-specific values of $2^n$ (refer to table X2 in appendix), where the *n* is determined by the GPU capacity. All the training and the testing tasks are performed on NVIDIA GeForce GTX 2080Ti graphics cards using the popular frameworks PyTorch(Paszke, et al., 2017) and the PyTorch pretrained models library. In comparison, the EfficientNet models are also trained by applying the identified AutoAugment strategy on ImageNet(Cubuk, et al., 2019) and by applying the General Augmentation strategy that composes only random flip and color jitter.

*2.4 Metrics for cost and performance evaluation*

The computational cost of the proposed strategy is evaluated by the search space size, defined as the order of magnitude for the number of possible augmentation policies(Cubuk, et al., 2020). The diagnostic performance of the proposed strategy is evaluated by the balanced accuracy (BACC) across the seven classes, equivalent to the average recall or sensitivity(Gessert, et al., 2018). As shown in figure 3, a multi-crop evaluation strategy is used for the generation of the final predictions, and the performance is generally better after averaging is applied [Gessert, et al., 2018]. Specifically, 16 region of interests (ROIs) with the size of 224 x 224 were equidistantly cropped from the upper left corner to the lower right corner of each unscaled image, and an average across all the predictions is used as a benchmark for final prediction.

The following metrics are used to evaluate the prediction performance on class $i$:

$$Precision_i = TP_i/(TP_i + FP_i) \quad (4)$$
$$Sensitivity_i = TP_i/(FN_i + TP_i) \quad (5)$$
$$Specificity_i = TN_i/(TN_i + FP_i) \quad (6)$$
$$Accuracy_i = (TN_i + TP_i)/(TN_i + FP_i + FN_i + TP_i) \quad (7)$$

where $TP_i$ is the number of true positive cases in class $i$; $FN_i$ is the number of false negative cases in class $i$; $TN_i$ is the number of true negative cases; $FP_i$ is number of false positive cases, all in class $i$.

The key metric of BACC for ISIC 2018 challenge is defined in equation 8, which is also used as the metric for our preliminary performance evaluation and hyperparameter tuning.

$$BACC = \frac{1}{7}\sum_{i=1}^{7} TP_i/(TP_i + FN_i) = \frac{1}{7}\sum_{i=1}^{7} Sensitivity_i \quad (8)$$

## 3. Results and Discussion

*3.1 Evaluation of augmentation strategies in 5-fold cross-validation*

Totally 12 sub-policies in the search space of Low-Cost-Augment are defined. Each sub-policy comes with 5 uniformly spaced probabilities ([0.1, 0.3, 0.5, 0.7, 0.9]) and random magnitudes, leading to a search space size of 12 x 6 = 60 possibilities. In comparison, the search space sizes of AutoAugment(Cubuk, et al., 2019), Fast AutoAugment (Lim, Kim, Kim, Kim, & Kim, 2019), Population Based Augmentation(Ho, et al., 2019) and RandAugment are $10^{32}$, $10^{32}$, $10^{61}$, and $10^2$ respectively. Notably, the proposed strategy greatly reduces the search space and thereby decreases the computational costs.

Table 2 - BACC of EfficientNet b0-b7 trained adopting the LCA strategy in different probabilities (P).

| CNNs | P=0.1 | P=0.3 | P=0.5 | P=0.7 | P=0.9 |
|---|---|---|---|---|---|
| **EfficientNet b0** | 0.881 ±0.025 | 0.883 ±0.014 | 0.881 ±0.015 | 0.874 ±0.021 | 0.875 ±0.024 |
| **EfficientNet b1** | 0.881 ±0.033 | 0.882 ±0.016 | 0.873 ±0.020 | 0.880 ±0.027 | 0.873 ±0.027 |
| **EfficientNet b2** | 0.879 ±0.019 | 0.883 ±0.013 | 0.881 ±0.016 | 0.876 ±0.024 | 0.880 ±0.029 |
| **EfficientNet b3** | 0.873 ±0.027 | 0.870 ±0.016 | 0.867 ±0.018 | 0.868 ±0.015 | 0.859 ±0.019 |
| **EfficientNet b4** | 0.878 ±0.021 | 0.876 ±0.013 | 0.874 ±0.019 | 0.871 ±0.020 | 0.873 ±0.014 |
| **EfficientNet b5** | 0.859 ±0.032 | 0.868 ±0.018 | 0.858 ±0.024 | 0.858 ±0.023 | 0.845 ±0.017 |
| **EfficientNet b6** | 0.871 ±0.016 | 0.866 ±0.012 | 0.857 ±0.033 | 0.862 ±0.033 | 0.856 ±0.033 |
| **EfficientNet b7** | 0.864 ±0.028 | 0.865 ±0.013 | 0.865 ±0.012 | 0.851 ±0.032 | 0.848 ±0.019 |

Table 2 shows the BACC performance of the trained EfficientNets after applying the proposed LCA strategy at various probabilities on the HAM10000 dataset. According to the table, the BACC values show a trend of first increasing and then decreasing from EfficientNets b0-b7, and the deviations fluctuate within ±0.033 in 5-fold cross-validation, no matter which augmentation strategy is used. These results imply that the most complex network is not necessarily suitable for the HAM10000 dataset of medium-capacity. It is also observed that the data augmentation strategy in the probability of 0.1 or 0.3 generally performs better than that of the other probabilities, as five out of the eight models yield the higher BACC values at the probability of 0.3.

The BACC performance of the LCA strategy at the probability of 0.3 is compared with other augmentation strategies, as shown in figure 4. According to the figure, the performance of the LCA strategy obviously exceeds both the General Augmentation and the searched AutoAugment strategies. These results indicate that the LCA strategy at the probability of 0.3 effectively reduces the overfitting risk and is more suitable for the HAM10000 dataset. It is also observed from the figure that the General Augmentation strategy performs better than the ImageNet-based AutoAugment strategy, indicating that the augmentation strategy obtained from one dataset cannot be effectively transferred to the other datasets.

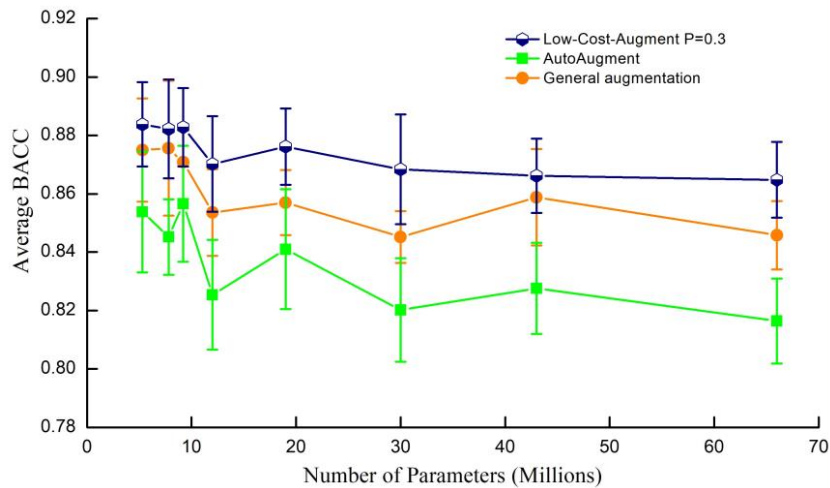

**Fig. 4 - BACC performance of EfficientNets trained adopting the best LCA strategy (n=5), ImageNet-based AutoAugment strategy (n=5), and General Augmentation strategy (n=5). The DCNNs from left to right on the X-axis correspond to EfficientNet b0-b7.**

*3.2 Performance validation*

EfficientNet b0-b7 are finely tuned by adopting the two-stage data augmentation strategy at the probability of 0.3 on the HAM10000 full dataset. The predicted results of the test dataset on the fine-tuned models at different epochs are uploaded to the official website of ISIC 2018 Challenge in order to obtain the BACC values and other performance metrics. Based on our experience, the BACC value for the models of over 30 epochs only fluctuate within a small range, therefore we collect the official test results in the range of 30~90 epochs with 5 epoch intervals.

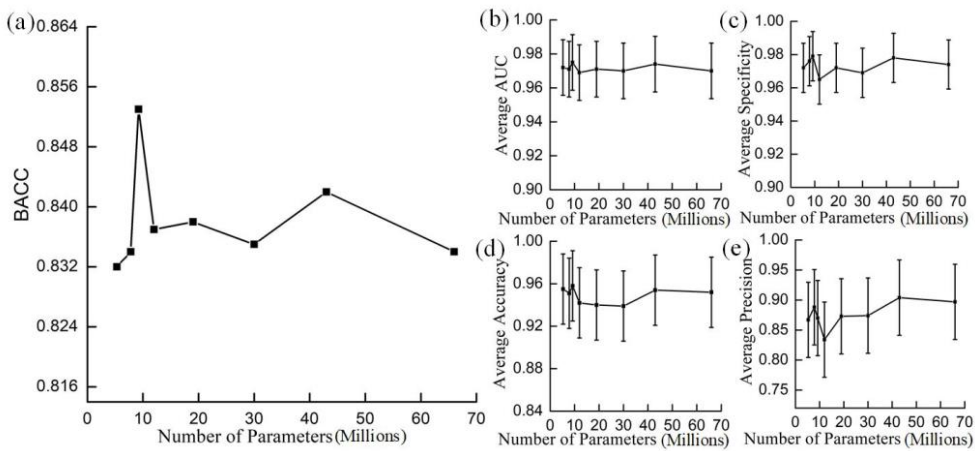

**Fig. 5 - (a) BACC, (b) average AUC (n=7), (c) average specificity (n=7), (d) average accuracy (n=7), and (e) average precision (n=7) of EfficientNets trained adopting the searched augmentation strategy. The DCNNs from left to right on the X-axis correspond to EfficientNet b0-b7.**

Figure 5 shows the best official performance of the EfficientNet models trained by the proposed LCA strategy. Although no obvious correlation is observed between the optimal BACC value and the corresponding parameters of different EfficientNet models, the EfficientNet b2 achieves the highest BACC value of 0.853, better than any other models. Similar trends are also observed in other metrics of the figure, such as the average area under the receiver operating characteristic curve(AUC), the average accuracy, the average specificity, and the average precision. Further analysis of the EfficientNet b2 performance in seven classes of HAM10000 (table X3 in appendix) indicates that the diagnostic specificity of the model is greater than 0.983 for all the classes, and the diagnostic accuracy is greater than 0.91 for all the classes except NV.

The performance of our augmentation strategy is also compared with those on the ISIC 2018 Challenge legacy leaderboard, as listed in Table 3. According to the table, using ensembling multiple models or using external datasets typically achieve better performance, such as those by Minjie, MetaOptima Technology Inc(Nozdryn-Plotnicki, et al., 2018), DAISYLab(Gessert, et al., 2018), Amirreza(Mahbod, et al., 2020), and IPM-HPC. In the case where only a single model is used without external database, the BACC reported on the legacy leaderboard[2] typically does not exceed 0.8, and the latest study on the live leaderboard[3] does not exceed 0.836. Therefore, the BACC performance of EfficientNet b2 using our augmentation strategy ranks the first on the channel of "single-model" and "no-external-data", even better than some of the ensembling models. It is also worthwhile to note that our proposed strategy achieves the melanoma diagnostic sensitivity superior to the ensembling models and those using external datasets.

**Table 3 - The performance of models on ISIC 2018 challenge legacy leaderboard[2] (rows 1–3), live leaderboard[3] (rows 4–11), and our proposed approach.**

| Team / authors | Use external data | Use ensemble models | BACC | Sensitivity for melanoma | Avg. AUC | Avg. specificity |
|---|---|---|---|---|---|---|
| **Nozdryn et al.** (Nozdryn-Plotnicki, et al., 2018) | Yes | Yes | 0.885 | 0.760 | 0.983 | 0.833 |
| **Gassert et al.** (Gessert, et al., 2018) | Yes | Yes | 0.856 | 0.801 | 0.987 | 0.984 |
| **Zhuang et al.** | No | Yes | 0.845 | 0.702 | 0.978 | 0.980 |
| **Minjie** | Yes | Yes | 0.895 | 0.778 | 0.982 | 0.981 |
| **Amirreza et al.** | Yes | Yes | 0.874 | 0.585 | 0.979 | 0.992 |
| **IPM_HPC.** | Yes | Yes | 0.866 | 0.830 | 0.976 | 0.976 |
| **Amirreza et al.** (Mahbod, et al., 2020) | Yes | Yes | 0.862 | - | 0.981 | - |
| **Mahbod et al.** | No | No | 0.836 | 0.719 | 0.975 | 0.982 |
| **ND Reddy et al.** | No | No | 0.735 | 0.544 | 0.945 | 0.968 |
| **Our approach.** | No | No | 0.853 | 0.789 | 0.975 | 0.973 |

*3.3 Discussion*

AI-based diagnostic techniques can be potentially used to not only relieve dermatologists and dermatopathologists from time-consuming or repetitive tasks but also provide expert dermatologic care to rural populations, underserved communities, and regions of limited resources(Evans & Whicher, 2018; Fine, 2016; Philipp Tschandl, et al., 2019). Inspired by recent advances in computing technology and DCNN, various AI-based skin disease classifiers have been developed. Although some of these classifiers have reported remarkable diagnostic accuracies equal or even superior to that of dermatology specialists, the outstanding performance is limited to the specific datasets and can hardly replicated in general clinical data with consistent accuracy(Goyal, Knackstedt, Yan, & Hassanpour, 2020). Moreover, the outperformed DCNNs typically assemble multiple models that require significant computational resources or use external training sets inaccessible in public domain, hindering their deployment in rural communities and regions of limited resources.

This project aims at developing a low-cost and high-performance data augmentation strategy that can be implemented in a low-complexity DCNN model for automatic skin cancer screening in rural communities. The proposed data augmentation strategy includes two consecutive stages of augmentation search and network search in a novel LCA search space. Compared with commonly used augmentation strategies such as AutoAugment(Cubuk, et al., 2019), Fast AutoAugment(Lim, et al., 2019), Population Based Augmentation(Ho, et al., 2019) and RandAugment, the size of LCA search space is only 60, representing a significant reduction of the search space and the computational costs. The performance of the proposed augmentation strategy is verified on the HAM10000 dataset using EfficientNet models. The best combination of the augmentation strategy and the DCNN model yields a BACC value of 0.853, ranking the first in the channel of "single-model & no-external-data" for task 3 of ISIC 2018 Challenge. This result is even better than those of many ensemble models reported on the leaderboard. In addition to skin cancer classification, the proposed data augmentation strategy can be applied to other medical datasets in order to facilitate the development and deployment of low-cost, high-performance and AI-based mobile devices for potentially automatic screening of many diseases in rural communities and regions of limited resources.

---

[2] https://challenge.isic-archive.com/leaderboards/2018 (TASK 3: LESION DIAGNOSIS)

[3] https://challenge.isic-archive.com/leaderboards/live (2018.3: LESION DIAGNOSIS)

Although the proposed augmentation strategy has a superior performance in the channel of "single-model & no-external-data", its BACC is still behind the best performance on the ISIC 2018 challenge leaderboards by 0.4. This gap is possibly due to insufficient representation of data augmentation or possible overfitting of a single DCNN on the HAM10000 dataset. Future efforts will be made to update the augmentation search space in order to incorporate more invariances, add more data augmentation methods, and alleviate the issue of under-represented patients in the HAM10000 dataset(Jiang, et al., 2020). Considering the possible over-fitting of EfficientNets, we may also integrate new structures such as drop block or adopt other state-of-the-art DCNNs for further improvement of the network performance.

## *4. Conclusion*

This paper proposes a low-cost and high-performance data augmentation strategy that can be potentially implemented in AI-based mobile devices for automatic screening of skin lesions in rural communities. The proposed strategy includes two consecutive stages of augmentation searching and network searching in a novel LCA search space comprising 12 sub-policies. By treating augmentation search as a discrete optimization problem, we have reduced the search space size by up to several orders of magnitude, leading to the reduced computing resources and the shortened search time. The efficiency of the proposed data augmentation strategy is verified on the HAM10000 dataset using EfficientNets as baseline. The results indicate that the LCA space at a probability of 0.3 in combination with EfficientNet b2 yields the best BACC of 0.853, outperforming the other published models in the channel of "single-model && no-external-database". Further comparison of classification performance using different parameter sizes and different models implies that the increased network complexity may not result in the improved performance on the HAM10000 dataset. The proposed data augmentation strategy will be implemented in a low-cost, portable and AI-based mobile devices for skin cancer screening in rural communities. This strategy can be also implemented in other clinical disciplines for early screening and automatic diagnosis of many other diseases in areas of limited resources.

## CRediT authorship contribution statement

## Acknowledgements

The authors are grateful for the clinical inputs and advices by Dr. Siping Zhang and Dr. Faxin Jiang (Department of Dermatology at the First Affiliated Hospital of USTC) and Dr. Benjamin Kaffenberger (Division of Dermatology, The Ohio State University Wexner Medical Center). This research did not receive any specific grant from funding agencies in the public, commercial, or not-for-profit sectors.

## Declaration of Competing Interest

The authors declare that there are no conflicts of interest related to this article.

## Appendix

**Table X1 - List of all image transformations used in the Low-Cost-Augment. List of all image transformations used in the Low-Cost-Augment. Additionally, the values of magnitude that can be applied by the controller during the search for each operation are shown in the second column, and they refer to references(Ho, et al., 2019; Zoph, et al., 2019). Some transformations do not use the magnitude information (e.g. Invert and Equalize).**

| Operation name | Range of magnitude | Description |
|---|---|---|
| Sample_Pairing | [0, 0.4] | Linearly add the image with another image (randomly selected from the same batch) with weight magnitude, without changing the label. |
| Gaussian_noise | [0, 0.4] | Add random Gaussian noise to the image with rate magnitude. |
| SolarizeAdd | [1, 110] | For each pixel in the image that is less than 128, add an additional amount to it decided by the magnitude. |
| Color | [0.1, 1.9] | Adjust the color balance of the image, in a manner similar to the controls on a color TV set. A magnitude=0 gives a black & white image, whereas magnitude=1 gives the original image. |
| Contrast | [0.1, 1.9] | Control the contrast of the image. A magnitude=0 gives a gray image, whereas magnitude=1 gives the original image |
| Brightness | [0.1, 1.9] | Adjust the brightness of the image. A magnitude=0 gives a black image, whereas magnitude=1 gives the original image. |
| Sharpness | [0.1, 1.9] | Adjust the sharpness of the image. A magnitude=0 gives a blurred image, whereas magnitude=1 gives the original image. |
| Color_shift | [-20, 20] | Add a random magnitude value to R, G, and B channels respectively |
| Equalize_YUV | | Equalize the histogram of each YUV channel after transferring image into YUV color spaces. |
| Equalize | | Equalize the histogram of each RGB channel of image respectively |
| Posterize | [4, 8] | Reduce the number of bits for each pixel to magnitude bits. |
| AutoContrast | | Maximize the image contrast, by making the darkest pixel black and lightest pixel white |
| Rotate | [-30, 30] | Rotate the image magnitude degrees. |
| Flip | | Flip image randomly in horizontal or vertical axis |
| Cutout | [0, 60] | Set a random square patch of side-length magnitude pixels to gray. |
| ShearX(Y) | [-0.3, 0.3] | Shear the image along the horizontal (vertical) axis with rate magnitude. |
| Scale | [0.6, 1.4] | Randomly scale the picture proportionally with rate magnitude. |

**Table X2 - Batch-size used during training for different EfficientNets**

| CNNs | Batch-size | CNNs | Batch-size | CNNs | Batch-size | CNNs | Batch-size |
|---|---|---|---|---|---|---|---|
| **EfficientNet b0** | 64 | **EfficientNet b2** | 64 | **EfficientNet b4** | 32 | **EfficientNet b6** | 16 |
| **EfficientNet b1** | 64 | **EfficientNet b3** | 32 | **EfficientNet b5** | 16 | **EfficientNet b7** | 16 |

Table X3 – The detailed performance of the best model on the official test data. (AUC: Area under the receiver operating characteristic curve).

| Category Metrics | Mean Value | Diagnosis Category | | | | | | |
|---|---|---|---|---|---|---|---|---|
| | | MEL | NV | BCC | AKIEC | BKL | DF | VASC |
| AUC | 0.975 | 0.943 | 0.970 | 0.989 | 0.988 | 0.968 | 0.972 | 0.995 |
| Average Precision | 0.870 | 0.785 | 0.981 | 0.897 | 0.826 | 0.873 | 0.813 | 0.915 |
| Accuracy | 0.958 | 0.925 | 0.902 | 0.980 | 0.979 | 0.938 | 0.990 | 0.991 |
| Sensitivity | 0.853 | 0.789 | 0.869 | 0.882 | 0.907 | 0.811 | 0.795 | 0.914 |
| Specificity | 0.979 | 0.958 | 0.962 | 0.993 | 0.981 | 0.968 | 0.996 | 0.993 |